# Neural network algorithm and its application in reactive distillation

Huihui Wang, Ruyang Mo

School of Chemical Engineering, East China University of Science and

Technology，Shanghai，200237

**Abstract:** Reactive distillation is a special distillation technology based on the coupling of chemical reaction and distillation. It has the characteristics of low energy consumption and high separation efficiency. However, because the combination of reaction and separation produces highly nonlinear robust behavior, the control and optimization of the reactive distillation process cannot use conventional methods, but must rely on neural network algorithms. This paper briefly describes the characteristics and research progress of reactive distillation technology and neural network algorithms, and summarizes the application of neural network algorithms in reactive distillation, aiming to provide reference for the development and innovation of industry technology.

**Keywords:** Reactive Distillation, Neural Network, BP Neural Network, RBF Neural Network

# 1 Introduction

Neural network, also called artificial neural network (ANN), is an artificially established dynamic system composed of multiple simple processing units, which are connected to each other in a certain way [1], and can process information by adjusting the relationship between a large number of nodes. ANN can accurately identify and learn the potential relationship between input and output, thereby predicting or optimizing system performance [2]. In addition, artificial neural networks have high self-learning ability and high fault tolerance, so they are widely used in various fields, such as medicine, national defense, transportation [3], machinery, finance, chemical industry, big data assimilation [4] and so on. For example, in the chemical industry, ANN applications include predictive models, spectrum analysis, structure and property prediction, signal processing, process control and optimization, quality control, quantitative prediction of physical properties, cluster analysis, expert systems, etc. There are many models of artificial neural networks. The commonly used ones are BP neural network and RBF neural network. They can be applied to the optimization control [5] and simulation of various chemical processes with multiple input parameters and multiple control objectives, such as reactive distillation and extraction distillation, supercritical extraction, etc.

Reactive distillation is different from ordinary distillation. It integrates reaction and distillation in one device, and the two processes occur at the same time and at the same place. It has the characteristics of low investment, low energy consumption, and high separation efficiency. In 1921, scientists first proposed the concept of reactive distillation, but it wasn't until 1980 that reactive distillation technology really got practical applications [6] and was incorporated into the future efficient and flexible (bio)chemical production process [7]. The reactive distillation process has the following characteristics: it can increase the conversion rate by removing products from the reaction zone; for chemical systems that tend to form azeotropes, the participating components can be removed by "reaction" to avoid azeotropes; for exothermic reactions, the reaction heat can be directly coupled to reduce the total heat required and avoid the occurrence of hot spots; the amount of catalyst used in the reaction is significantly reduced [7]. On the other hand, reactive distillation also has certain limitations. For example, the chemical reaction must occur in the liquid phase; the residence time of the reaction cannot be too long; in order to ensure the practical application effect, the reaction cannot be a strong endothermic reaction [8].

At present, the simulation of reactive distillation process includes steady state simulation [9] and unsteady state simulation. Due to the interaction of chemical reaction and rectification separation, in reactive distillation process there will be nonlinear phenomena and multi-stable phenomena, that is, one input (or output) corresponds to multiple outputs (or inputs). Higler *et al.* [10] developed a generic non-equilibrium (NEQ) model for packed reactive distillation columns. The important features of the model are the use of the Maxwell-Stefan equations for description of intraphase mass transfer and incorporation of a homotopy-like continuation method that allows for easy tracking of multiple steady states. Mekala *et al.* [11] simulated the rectification process of the esterification reaction of acetic acid and methanol using the equilibrium model. Experimental simulations have obtained the optimal conditions of acetic acid and methanol and the high conversion rate of acetic acid. The non-steady-state simulation of reactive distillation has a wide range of applications than steady-state distillation. It does not need to quote the number of mass transfer units and stage efficiency, and is suitable for homogeneous reactions.

Due to the strong interaction between chemical reaction and heat and mass transfer, the design of enhanced separation process is often complicated [12]. Moreover, the combination of reaction and separation has produced highly nonlinear robust behavior and making the control and optimization of reactive distillation process unable to adopt conventional methods. To deal with the nonlinear robust optimization problem, some researchers proposed a distributional robust game framework[13], others use neural networks in which the neurons are optimized interactively. The ANN has the function of nonlinear mode mapping, strong self-learning ability and high fault-tolerant ability. It can carry out fast and accurate optimization simulation under the complex relationship between input and output, which can be used for the optimal control of reactive distillation column. This article will briefly describe the characteristics and research progress of neural network algorithm and reactive distillation technology, and summarize the application of neural network algorithm in reactive distillation.

# 2 Neural network algorithm and its research progress

Artificial neural network is referred to as neural network. Artificial nerve is widely used in various fields due to its high self-learning ability and high fault tolerance. Its emergence has greatly improved the technology and production environment in various fields, and provided practical solutions to the problems that traditional technology is difficult to handle.

## 2.1 Principle of Neural Network

Artificial neural network is a non-linear, adaptive information processing system composed of a large number of interconnected processing units. It mimics the human brain neuron network for abstraction, and then establishes a certain mathematical model. By adjusting the relationship between a large number of nodes in the model, the information is processed [14].

## 2.2 Classification of neural network models

### 2.2.1 BP neural network

BP (Back Propagation) type neural network, namely error back propagation artificial neural network, was proposed by Rumelhart, Hinton and Williams [15]. It consists of a large number of nodes and their interconnections, including input layer, hidden layer and output layer. It is a multi-layer feedforward neural network trained according to the back propagation algorithm [16].

BP neural network is different from linear regression model, it has no strict requirements on data distribution [17]. The characteristic of BP algorithm is to estimate the error of the direct leading layer of the output layer by using the error after output, and then use this error to estimate the error of the previous layer, so that the error estimates of all other layers can be obtained by the backward propagation of one layer.

In this way, the error shown by the output layer is transferred to the input layer of the network step by step along the direction opposite to the input transmission.

BP neural network is one of the most widely used neural networks at present, widely used in finance, accounting, management science, e-commerce and computer science. At the same time, some studies have confirmed the advantages of BP neural network in prediction. For example, Lee *et al.* [18] found that the prediction performance of the BP neural network model is better than the linear regression model. Due to the poor generalization ability and approximation ability of BP neural network and the existence of "overfitting" problem, the predictive ability of related models is not optimal.

### 2.2.2 RBF Neural Network

In the human cerebral cortex, there are locally regulated and overlapping receptive fields. According to this characteristic of the human brain, Moody and Darken proposed a radial basis basis (RBF) neural network, which is an abstraction and simplification of the human brain neural network system [19]. The RBF neural network is a three-layer feedforward neural network, which includes an input layer, a hidden layer, and an output layer. The transformation from the input node to the hidden layer node is nonlinear, and the transformation from the hidden layer node to the output node is linear, so any continuous function can be approximated with arbitrary precision, which is very suitable for nonlinear dynamic modeling [20].

Compared with BP neural network, RBFNN not only has superior clustering and classification capabilities, but also has better generalization capabilities and higher approximation accuracy [21]. Due to its simple learning algorithm and network structure, RBFNN has fast convergence and can uniformly approximate any continuous function to achieve the expected accuracy. Therefore, it is particularly suitable for the control of nonlinear and time-varying dynamic systems, but the uncertainty and parameter changes require additional attention [22].

## 2.4 Application of Neural Network

In the case of complex relationships between input and output, neural networks are believed to be able to produce accurate estimates because they map in a non-linear manner. Therefore, neural networks are widely used in various fields, such as mechanical fault detection[23], classification[24], image analysis[25], speech generation[26], detection of presentation attacks [27], prediction problems[28], vehicle tracking[29], and so on.

In terms of prediction, Kumar *et al.* [30] compared neural network method with traditional function point method, usecase method and COCOMO method, and the results showed that the prediction result of neural network method is more accurate and better. Song *et al.* [31] established an artificial neural network model (ANN-GCS) for $CO_2$ geological storage in salt-bearing formations and tested the validity of the model. In terms of mechanical fault detection, Xu *et al.* [32] and others pointed out the advantages of fuzzy neural network technology in equipment fault diagnosis, which can make up for the shortcomings of using fuzzy theory and neural network alone, and has high diagnostic accuracy, so it has a good application prospect in the field of rotating machinery fault diagnosis. In terms of optimization design, Dayev et al. [33] replaced the traditional flow coefficient equation with an artificial neural network, which can quickly change the synaptic coefficient of the network, and modify the flow coefficient model through additional input to the network. Kelley et al. [34] proposed a method to directly measure the neutron decay time, namely ANN. Using this method, the decay can be clearly shown, instead of calculating the neutron decay time based on the half-life results like other experiments. In terms of classification, Sang et al. [35] uses genetic algorithm and BP neural network optimized support vector machine method to classify enterprises. The results show that the classification accuracy of the support vector machine optimized by the genetic algorithm is 76.27%, and the classification accuracy of the BP neural network method is 89.83%. This research can provide theoretical support for banks to carry out related businesses. In terms of image analysis, Rabiej et al. [36] used neural network algorithms to identify the types of polymers based on the shape of their diffraction curves. It can be found that neural network algorithms can also be used in other methods with appropriate modifications, especially infrared spectroscopy and Raman Spectrum, etc. can be analyzed and compared with this algorithm.

Although neural networks have many advantages and are widely used in various fields. However, neural networks tend to use only the training data required for their tasks, and often ignore additional useful information. Especially in safety-related applications, neural networks should check their decision-making priorities. Seibold *et al.* [37] pointed out that neural networks that only focus on certain parts of the image have huge flaws in the lack of robustness against target attacks.

In addition, neural networks are also widely used in reactive distillation in the chemical industry.

# 3 Application research progress of reactive distillation technology

## 3.1 Application of reactive distillation in esterification reaction

An *et al.* [38] used reactive distillation to produce methyl acetate (MeAc) in a reactive dividing wall column (RDWC), and used chemical simulators Aspen Plus and Aspen Dynamics to study the design and control of RDWC. The results show that the RDWC design can save about 7.7% of energy consumption, 8.3% of operating costs and 15.5% of capital investment. In addition, the use of reactive distillation to synthesize MeAc improves the overall conversion rate and selectivity, and reduces raw material waste.

Propylene carbonate (PC) reacts with methanol to produce transesterified dimethyl carbonate (DMC), which will produce unfavorable chemical equilibrium and complex thermodynamic behavior. Reactive distillateddion is considered to be a potential candidate method to improve the efficiency of the chemical equilibrium restriction system and the product yield. Holtbruegge *et al.* [39] conducted experiments to simultaneously produce DMC in a pilot-scale reactive distillation tower, and found that the use of catalytic distillation to synthesize DMC can significantly increase the conversion rate of the reactant PC.

## 3.2 Application of reactive distillation in ether formation reaction

In the production of ethers, scientific researchers often prefer reactive distillation technology. Kiatkittipong *et al.* [40] used tert-butanol and glycerol etherification reaction to rectify the synthesis of glycerol ether and performed a series of simulations. Yamaki *et al.* [41] synthesized tert-amyl methyl ether (TAME) by reactive distillation and found that the reboiler load required to obtain high-purity TAME increased with the increase in reflux ratio.

## 3.3 Application of reactive distillation in hydrolysis reaction

Lactic acid (LA) is an important natural organic acid with many applications in the food, chemical and pharmaceutical industries. LA can be produced by fermentation or chemical synthesis. However, due to the strong affinity of LA to water, low volatility and self-polymerization tendency, it is difficult to obtain LA with high purity and thermal stability. Andrés *et al.* [42] used fermentation broth as a raw material for esterification, and then hydrolyzed lactate in a reactive distillation tower. This method is an effective lactic acid purification process. In order to study the hydrolysis reaction and rectification process of methyl lactate, Andrés conducted a lot of analysis and design, and finally proposed a dual temperature control structure for the optimized reaction rectification tower. The results show that this structure has a good control effect on the hydrolysis system. Li *et al.* [43] also proposed a potentially sustainable reactive distillation (RD) process to enhance the hydrolysis of 2-methyl-1,3-dioxolane (2MD). The designed RD process can make the traditional hydrolysis process develop towards a 2MD conversion rate greater than 99.9% without adding excessive water.

### 3.4 Application of reactive distillation in other reactions

In addition to the above reactions, reactive distillation techniques are also commonly used in other reactions. Belaissaoui *et al.*[44] extended the feasibility analysis and preliminary design of the reactive distillation tower to the system involving gas phase chemical reactions, and successfully applied it to hydrogen production by high vapor phase decomposition.

Ethanolamine has been proven to be an extremely versatile intermediate in many chemical products, such as emulsifiers, corrosion inhibitors, and agricultural chemicals. In order to determine the application potential of the RD process in ethanolamine production, Liu *et al.* [45] compared the total annual cost (TAC) of the RD process and the traditional plug flow reactor process. The results showed that TAC was reduced by 39.1%. Therefore, the RD process has broad application prospects in ethanolamine production.

# 4 Application research of neural network in reactive distillation

Reactive distillation has been widely used due to its advantages of low investment, low energy consumption and high separation efficiency. However, reactive distillation is a special distillation process and its simulation method is more complicated, so this deficiency exists to a certain extent. Therefore, it is necessary to study a simple and efficient simulation method to reduce its investment, energy consumption, and improve efficiency.

Shi *et al.* [19] prepared dimethyl carbonate (DMC) in a catalytic distillation column by transesterification reaction between propylene carbonate (PC) and methanol, and established a functional relationship model between propylene carbonate conversion and operating conditions such as pressure, reflux ratio, feed mole and feed rate by using RBF neural network. The genetic algorithm is used to find the optimal operating conditions when the conversion rate of propylene carbonate is maximized. After 150 generations of evolution, the experimental operating pressure is 0.5MPa, the reflux ratio is 9, and the feed molar ratio (methanol/PC) is 9: 1. The processing capacity is $0.1187m^3$ liquid PC/h [$m^3$ catalyst], and the optimal value of the propylene carbonate conversion rate at the output of the neural network model is 45.79%. It can be concluded that the use of neural network in reactive distillation can achieve better simulation of multi-factor and multi-variable nonlinear models, and then the combination of neural network and genetic algorithm can optimize the obtained model on a global scale, thereby obtaining the best condition.

In reactive distillation, in order to achieve energy efficiency while meeting product quality constraints, optimizing distillation tower operations is also crucial. Most of the mechanical models of distillation systems assume the equilibrium of each stage, such models deviate from reality and cannot be truly reproduced. For this reason, Osuolale *et al.* [46] assumed the non-equilibrium stage. However, the non-equilibrium model involves a large number of variables. As a result, the differential equation of the distillation model may exhibit high differential exponents and may produce rigid dynamics. It involves a large number of differential equations and algebraic equations and a large number of model parameters, which requires a lot of calculations and is not suitable real-time optimization. The artificial neural network model (ANN) can learn complex functional relationships from the input and output data of the system, and its calculation is less, which is suitable for real-time optimization. Therefore, Osuolale *et al.* proposed a neural network-based energy efficiency modeling and optimization strategy for distillation towers based on the second law of thermodynamics. The results show that the ANN network can accurately model the exergy efficiency from the process operation data of the distillation column. Then use the ANN network model to obtain the optimal distillation operating conditions to maximize the energy performance of the distillation system while maintaining product quality and output. They also proposed a reliable strategy for the promotion of an improved prediction model based on a bootstrap aggregate neural network (BANN). BANN improves the accuracy of model predictions and also provides confidence intervals for model predictions. Exergy analysis is an effective method to determine the energy efficiency of the process. The application of two binary systems and one multi-component system shows that this method can significantly improve the exergy efficiency of the distillation column. After optimization, the exergy efficiency of the methanol-water system and benzene-toluene system increased by 11.2% and 1.8% respectively, and the consumption of public works was reduced to 8.2% (methanol-water) and 28.2% (benzene-toluene) respectively. The exergy efficiency of the multi-component system is increased to 32.4%, which does not incur any additional capital costs. The improved multi-component system (exergy) increased by about 31%, but increased additional costs. The advantages of generating these additional costs can be weighed and an informed decision can be made. Modeling and optimization based on ANN and BANN models are helpful to the decision-making of energy-saving operation and control of distillation towers.

The synthesis of methyl tert-amyl ether (TAME) by reactive distillation is a complicated nonlinear process. Sharma *et al.* [47] used three different control strategies: conventional PID control, model predictive control (MPC) and neural network predictive control (NNPC) to control the reactive distillation tower. Through the verification of various performance indicators, it is found that NNPC and MPC have smoother and better control performance than PID controllers when the set value of methanol feed flow changes and the load changes by $\pm 10\%$. This fully shows that the neural network can handle complex nonlinear problems and can reduce the workload required for controller model development.

Tun *et al.* [48] combined the GA application with the process simulator and developed a design to optimize the system's reactive distillation process. Then, the genetic algorithm was improved. Due to the randomness of the genetic algorithm, it was initialized with different random seeds, and the optimization operation was repeated 6 times to expand the search results to design a better process from the perspective of dynamic operation and control.

Rani *et al.* [49] developed a soft-sensing technique for rectification tower neural network control. The main goal of this work is to accurately measure and maintain the quality of the distillate in the shortest possible time in the presence of interference. For the successful control of the distillation process, online measurement of product components is very important. The measurement of the component analyzer has an abnormal time delay, so the product quality is affected. In addition, the use of analyzers requires high capital costs, frequent calibration and regular maintenance. Therefore, the composition analyzer is not an effective solution for accurate online composition measurement. In order to overcome the difficulties encountered in component measurement, inference or soft measurement techniques have recently been developed as viable alternatives to hardware sensors. Adaptive linear neuron (ADALINE) can be used to infer fraction composition based on temperature measurement. Rani et al. used adaptive linear networks and LM-based neural networks to design a soft sensor that estimates product components from the temperature distribution during distillation. The results show that ADALINE soft measurement is better than LM soft measurement in terms of accuracy, training time and storage space required for training. Using the developed soft sensor, two inference controllers, LMIC and ADIC, are designed to control the fraction composition of the multi-component distillation process. The control performance of ADIC is better than that of LMIC. ADALINE soft measurement has further improved its performance through online training, enabling it to adapt to changes in process input variables. Due to the two main defects of ANN: overfitting and difficulty in determining the optimal number of hidden points. In response to this problem, Yan[50] proposed a hybrid artificial neural network (HANN) that combines BP and partial least square regression (PLSR). First, HANN selects a single hidden layer network composed of input layer, single hidden layer and output layer. Determine the number of hidden layer neurons according to the number of model samples and the number of neural network parameters. Secondly, use the BP network to train the neural network, and then use the hidden layer to perform nonlinear transformations on the independent variables. Thirdly, by using the correlation between the nonlinear transformation variables of the inverse function of the activation function of the output layer nodes, the optimal relationship model of the nonlinear transformation variables expected by the input nodes of the output layer is obtained. Therefore, the HANN model was developed. In addition, due to the existence of many non-linear influencing factors and the significant correlation between the factors, HANN has developed a soft measurement software for naphtha dry point and the most important intermediate product in the oxidation reaction of paraxylene (ie 4-Carboxybenzaldehyde) concentration soft meter. Two application results show that HANN overcomes over-fitting and is robust. In the two applications, the squared relative errors of the predictions of the two optimal HANN models are smaller than the squared relative errors of the two optimal ANN models, and the average squared relative errors of the predictions of HANN are also lower than ANN.

As we all know, neural networks have flexible function approximation capabilities. The application of RBF networks in nonlinear systems has been widely discussed. Usually RBF networks are used for single-input single-output systems, but they also provide the possibility for larger systems. The training speed of RBF networks is usually faster than that of feedforward perceptron networks. Säynätjoki *et al.* [51] discussed the application of RBF network in the identification of the nonlinear multi-input multi-output model of the pilot-scale reactive distillation column. First, the reactive distillation process of methyl tert-amyl ether produced by the reaction of methanol and isopentene was studied. During the pilot plant test, identification data was recorded. Three different operating areas of the process were tested. The results show that the model can predict different regions well.

Mohammed *et al.* [52] used Neural Network Model Predictive Control (NNMPC) to deal with the optimal strategy tracking problem determined by the dynamic optimization strategy of the batch reaction distillation column. The multi-layer feedforward neural network model and estimator are established and applied to the model predictive control algorithm. The results show that NNMPC has a good control performance for the setpoint tracking problem. The robustness of NNMPC to object/model mismatch is studied and compared with the traditional proportional controller (P). Experimental results show that NNMPC has better control performance than P controller in all cases.

# 5 Conclusion

As a special distillation technology based on the coupling of chemical reaction and distillation, reactive distillation has been widely used in various fields due to its advantages of low energy consumption, high separation efficiency, low investment, and high purity. However, due to the complexity of the simulation method of reactive distillation, the combination of reaction and separation has produced highly nonlinear robust behavior and other deficiencies to a certain extent, the control and optimization of the reactive distillation process cannot be controlled and optimized by conventional methods. This article discusses the application of neural network algorithm in reactive distillation. Artificial neural network has high self-learning ability and high fault tolerance, and can accurately identify and learn the potential relationship between input and output. In addition, because artificial neural networks do not require the assumptions of mathematical models, it is considered an effective alternative to traditional statistical techniques in function approximation and data fitting. Due to these advantages of artificial neural network, it can find the best reaction conditions in reactive distillation and obtain the maximum conversion rate of reactants, which can save costs and raw materials, and make the separation purity and efficiency of the reaction high. That is, to a certain extent, the neural network can optimize the reactive distillation technology. The research progress of neural network algorithm in reactive distillation process summarized in this paper in recent years, hope to provide reference for the development and innovation of industry technology.

artificial neural network for the prediction of catalytic performances, Applied Catalysis, 2013, 466(10): 142-152